\documentclass[sigconf]{acmart}
\AtBeginDocument{%
  }

\setcopyright{acmlicensed}
\copyrightyear{2026}
\acmYear{2026}
\acmDOI{XXXXXXX.XXXXXXX}
\acmConference[KDD '26]{Proceedings of the 32nd ACM SIGKDD Conference on Knowledge Discovery and Data Mining}{August 3--7,
  2026}{Washington, DC, USA}
\acmISBN{978-1-4503-XXXX-X/2018/06}

\begin{document}

\title{The Sim-to-Real Gap of Foundation Model Agents: \\A Unified MDP Perspective}

\author{Xiaoou Liu$^*$, Tiejin Chen$^*$, Weibo Li, Xiyang Hu, Hua Wei}
\email{{xiaoouli,tiejin,weiboli,xiyanghu,hua.wei}@asu.edu}
\affiliation{%
  \institution{Arizona State University}
  \city{Tempe}
  \state{Arizona}
  \country{USA}
}

\renewcommand{\shortauthors}{Authors et al.}

\begin{abstract}
Foundation model agents are increasingly deployed for real-world decision-making, but suffer from the sim-to-real gap. While robotics and classical control have mature frameworks to address this gap, the foundation model community is treating agent robustness as an entirely novel phenomenon. Our paper proposes formalizing the foundation model agent evaluation and training gap as a classical sim-to-real problem structured entirely around the four elements of a Markov Decision Process, including Observation, Action, Transition, and Reward. In this paper, we set a comprehensive research agenda that translates classical discrepancies into the foundation model domain and advocates for adopting established solutions like domain randomization. We provide concrete examples, such as a multilingual tool calling to demonstrate how severe observation space gaps lead to operationally invalid actions despite correct semantic intent. Ultimately, this agenda aims to drive a paradigm shift, yielding a unified vocabulary and standardized stress test benchmarks to foster a new generation of highly trustworthy agents for reliable real-world applications.
\end{abstract}


\keywords{Sim-to-real, Reinforcement Learning, LLM Agents, MDP}



\maketitle

\section{Introduction}

\begin{figure*}[t]
    \centering
    \includegraphics[width=\linewidth]{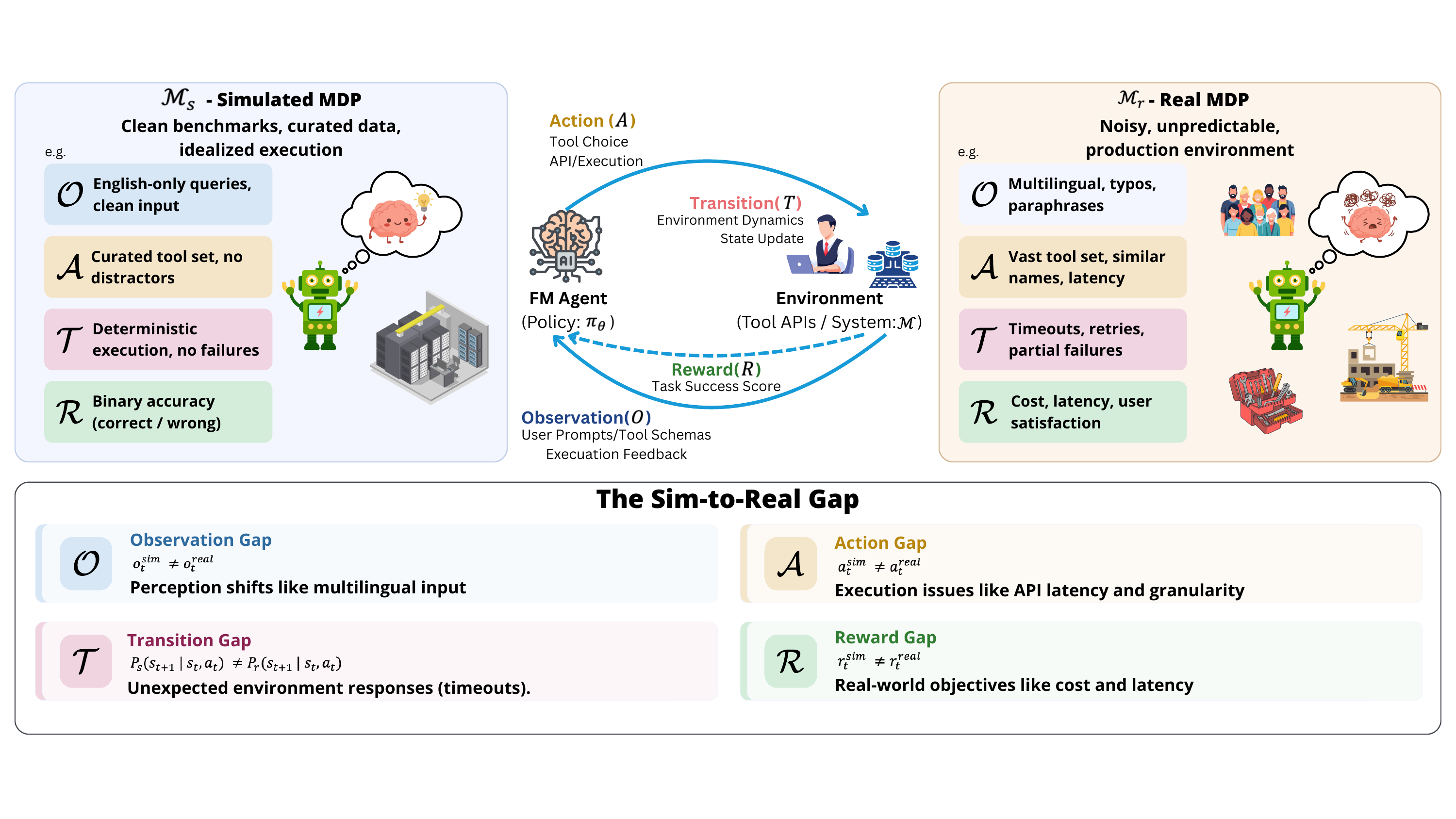}
    \vspace{-16mm}
    \caption{
Landscape view of sim-to-real transfer for foundation model agent systems under the MDP decomposition.
A policy $\pi$ trained in the simulated MDP $\mathcal{M}_s$ (left) is deployed without modification in the real MDP $\mathcal{M}_r$ (right).
Performance degradation arises from discrepancies in four channels—observation, action, transition, and reward—represented as a central “gap wall.”
Each channel corresponds to distinct real-world shifts (e.g., multilingual observation mismatch, action-space distractors, stochastic tool execution, cost-aware reward signals) and can be stress-tested via controlled perturbations.
The same decomposition aligns directly with classical sim-to-real remedies, enabling unified benchmarking and training across $\mathcal{O,A,R,T}$. 
}
    \label{fig:sim2real_landscape_framework}
\end{figure*}

With the exploration of foundation models~\cite{singh2025openai,grattafiori2024llama,qwenteam2025qwen3next}, foundation model (FM) agents and reinforcement learning (RL) policies are increasingly deployed for real-world decision-making, spanning tool-augmented assistants~\cite{li2023camel}, robotic control~\cite{kim2024survey}, and autonomous systems~\cite{yao2025instructional}. While training typically occurs on clean, curated benchmarks where data is abundant, and failures are safe, deployment exposes policies to distribution shifts, noisy inputs, unpredictable execution constraints, and stochastic transitions. Consequently, performance on the leaderboard does not equate to real-world deployment reliability~\cite{da2025survey}. This simulation-to-reality (sim-to-real) gap is extensively studied in robotics and classical control, which boast mature frameworks to address such discrepancies~\cite{da2024prompt,hanna2017grounded}. However, as FMs become the backbone of agent systems, the community is currently reinventing the wheel by treating agent robustness as an entirely novel phenomenon rather than a known deployment gap~\cite{zhu2025llm,zhou2026simulation}. \textit{We should correct this isolated perspective and start to solve the sim-to-real gap in FM agents now} because FM policies are rapidly moving from safe simulators into unpredictable production environments where ignoring established sim-to-real frameworks leaves critical vulnerabilities unmeasured and causes severe real-world failures.

In this paper, \textit{our big and bold idea is to formalize the FM agent evaluation and training gap as a classical sim-to-real problem structured entirely around the four elements of a Markov Decision Process, including Observation, Action, Transition, and Reward}. Recent evaluations, such as tool-use perturbation benchmarks and studies on multilingual tool calling, are inadvertently rediscovering these exact gaps~\cite{rabinovich2025robustness,wang2026agentnoisebench}, but they completely lack a unifying language. We advocate that the agent community should adopt the same family of classical solutions, including domain randomization~\cite{bellemare2016increasing,wiltzer2024action} and grounded action transformation~\cite{hanna2017grounded,lin2025revealing,karnan2020reinforced,da2023uncertainty}. Establishing this unified formulation is critical at this moment to prevent fragmented research efforts and to accurately measure deployment readiness.

To push the frontier and offer a bold approach to this sim-to-real gap, we operationalize the classical Markov Decision Process view across the subsequent sections of this paper. In detail, we build upon recent surveys of simulation to reality methods in reinforcement learning~\cite{da2025survey, zhao2020sim} that organize techniques by the four elements of the Markov Decision Process (MDP), including observation ($\mathcal{O}$), action  ($\mathcal{A}$), transition  ($\mathcal{T}$), and reward  ($\mathcal{R}$). We review these four gaps in traditional reinforcement learning in Section~2, including their typical causes and classical mitigation techniques, and then dedicate Section~3 to translating these classical discrepancies into the FM domain. Within this section, we provide concrete examples for each component and propose how benchmarking paradigms can be designed to systematically expose these vulnerabilities. To demonstrate that such gaps genuinely exist in practice, we incorporate recent studies on multilingual tool calling~\cite{luo2025multilingual} which highlight a severe observation space gap where models correctly understand user intent and select the appropriate tool, yet produce parameter values in the user language that violate strict execution level conventions, leading to operationally invalid tool calls. Finally, Section 4 outlines concrete research directions to harden agents against these gaps.

\textit{If all of our ideas succeed, the outcome would look like a fundamental paradigm shift where the field shares a unified vocabulary for vulnerabilities and universally adopts standardized stress test benchmarks}. This collective shift will directly yield a new generation of highly trustworthy agents that intrinsically maintain performance despite severe multilingual inputs and transition perturbations and cost-aware reward constraints. Researchers building foundation models will gain a structured methodology for robustness testing while practitioners deploying autonomous systems will acquire a reliable blueprint for safe production environments. These specific groups and their deployment challenges represent the core focus of the KDD community. Therefore, this agenda directly advances KDD thematic topics by establishing a rigorous foundation for Trustworthy and Responsible Data Science and shaping Modern AI for reliable real world Applications.

\section{Problem Formulation and Classical Solutions}

RL is commonly formulated as a discounted Markov decision process (MDP)~\cite{feinberg2012handbook} $\mathcal{M} = (\mathcal{S}, \mathcal{A}, \mathcal{T}, \mathcal{R}, \gamma)$, where $\mathcal{S}$ is the state space, $\mathcal{A}$ is the action space, $\mathcal{T}(s_{t+1}\mid s_t, a_t)$ is the transition dynamics, $\mathcal{R}$ is the reward function, and $\gamma \in [0,1)$ is the discount factor. The objective is to learn a policy $\pi$ that maximizes the expected discounted return. In sim-to-real settings, the policy is trained in a simulated MDP $\mathcal{M}_s$ and deployed in a real-world MDP $\mathcal{M}_r$. Following prior work~\cite{da2025survey}, the \emph{sim-to-real gap} of policy $\pi$ can be defined as
$G(\pi) := \psi_s(\pi) - \psi_r(\pi)$, where $\psi_s(\cdot)$ and $\psi_r(\cdot)$ denote the same evaluation metric computed in simulation and in the real environment, respectively. From an MDP perspective, the source of $G(\pi)$ can be attributed to discrepancies in the core elements of the process: observation, action, transition, and reward.

\vspace{1mm}
\noindent$\bullet$~\textbf{Observation gap.} In sim-to-real RL, the agent's perceived observation in simulation often differs from that in the real world, i.e., $o_t^{s} \neq o_t^{r}$, which can induce a systematic performance drop after deployment. There are two common sources of the observation gap. 
(1)~\emph{Perception completeness}: simulators may provide ideal or fully-observed signals (e.g., $o_t^{s}=s_t^{s}$), whereas real observations are partial, noisy, delayed, or subject to occlusion.
(2)~\emph{Representation mismatch}: differences in sensor modalities, resolution, calibration, and encoding create a distribution shift in the observation space, even when the underlying task dynamics are similar.
\textbf{Techniques to mitigate} the observation gap include the following.
\emph{(1)~Domain Randomization} randomizes visual and sensing parameters in simulation to encourage invariance and improve out-of-distribution generalization~\cite{tobin2017domain}.
\emph{(2)~Domain Adaptation} aligns simulated and real observation feature distributions, for example via adversarial objectives~\cite{bousmalis2017unsupervised} or embedding alignment~\cite{park2021sim}, to reduce cross-domain discrepancy.
\emph{(3)~Sensor Fusion} combines complementary modalities to reduce reliance on any single biased channel, thereby improving robustness under real-world sensing imperfections.

\vspace{1mm}
\noindent$\bullet$~\textbf{Action gap.} In sim-to-real RL, actions that are valid and effective in simulation may not translate faithfully to the real world.
There are three common sources of the action gap: \emph{(1)~Action granularity}: simulators often use discretized or simplified action spaces and assume near-perfect execution, while real control is continuous, fine-grained, and constrained by low-level actuation limits;
\emph{(2)~Execution uncertainty}: real actuation is stochastic and imperfect, where intended actions can be perturbed or experience unexpected scaling in magnitude~\cite{bellemare2016increasing,wiltzer2024action};
\emph{(3)~System delay}: real actuators and APIs introduce latency and jitter, so the effective action may be applied at $t+\delta$ rather than immediately as assumed in simulation.
\textbf{Techniques to mitigate} the action gap include the following.
\emph{(1)~Action Shielding} projects or filters proposed actions to ensure feasibility and safety before execution.
\emph{(2)~Delay-aware Control} explicitly models latency (e.g., constant- or random-delay MDP variants) and trains policies robust to delayed actuation.
\emph{(3)~Robustification to Actuation Uncertainty} uses action perturbations/noise injection or robust RL objectives, improving stability under perturbed actions and action-scale shifts~\cite{tan2020robustifying,liu2024robust}.

\vspace{1mm}
\noindent$\bullet$~\textbf{Transition gap.}
Sim-to-real transfer often suffers from \emph{transition gaps}, where the next-state dynamics in simulation diverge from those in the real environment~\cite{hanna2017grounded,lin2025revealing}.
Such gaps arise from inaccurate or incomplete modeling of real dynamics.
\textbf{Techniques to mitigate} the transition gap include the following.
\emph{(1)~Transition-level Domain Randomization} perturbs dynamics parameters to train policies robust to model error~\cite{valassakis2020crossing,mehta2020active}.
\emph{(2)~Grounding Methods}~\cite{hanna2017grounded,desai2020stochastic,karnan2020reinforced,desai2020imitation} learn a transformation to map simulated transitions to real transitions~\cite{da2023uncertainty,karnan2020reinforced}.
\emph{(3)~Distributionally Robust Learning} optimizes policies to perform well under unknown-but-bounded transition shifts~\cite{smirnova2019distributionally}.

\vspace{1mm}
\noindent$\bullet$~\textbf{Reward gap.}
In sim-to-real RL, the reward specified in simulation may not faithfully reflect the real-world objective due to incomplete task modeling or cascading effects of observation, action, and transition mismatches~\cite{li2023mind}.
For example, delayed or mis-executed actions can alter realized outcomes and therefore change the reward received~\cite{kim2026cost}.
\textbf{Techniques to mitigate} the reward gap include the following.
\emph{(1)~Reward shaping} provides denser and more informative feedback while preserving the optimal policy under suitable conditions, as in potential-based shaping~\cite{badnava2023potential-based}.
Reward shaping can also incorporate structured priors such as automaton-guided shaping to better handle sparse objectives~\cite{velasquez2021dynamic}.
\emph{(2)~Reward augmentation} uses limited real-environment data to refine or supplement return signals, for instance by matching sim--real trajectory distributions or by augmenting returns for return-conditioned learning. This improves transferred policy performance under data scarcity~\cite{guo2024rewardaug}.

\begin{table*}[t]
\centering
\caption{Connecting traditional sim-to-real gap in RL to their analogues in FM agent, organized by the four MDP elements.}
\label{tab:traditional-fm-connection}
\vspace{-3mm}
\small
\begin{tabular}{@{}p{1.5cm}p{4.2cm}p{11.3cm}@{}}
\toprule
\textbf{MDP gap} & \textbf{Traditional Sim-to-real} & \textbf{Foundation Model Agents} \\
\midrule
Observation & 
\begin{tabular}[c]{@{}l@{}}
Representation mismatch~\cite{tobin2017domain}; \\
Observation distribution shift~\cite{park2021sim,bousmalis2017unsupervised}; \\
Perception incompleteness~\cite{bohez2017sensor};

\end{tabular}
& \begin{tabular}[c]{@{}l@{}}
\textbf{Observation perturbations}: typos, query/tool/param paraphrase~\cite{rabinovich2025robustness}. \\
\textbf{Distribution shift}: multilingual user input, language--execution mismatch (param.\ value \\ language mismatch)~\cite{luo2025multilingual}. Multi-modal extension~\cite{yu2024lang4sim2real}. \\

\end{tabular}
\\
\midrule
Action
& 
\begin{tabular}[c]{@{}l@{}}
Action granularity~\cite{abbas2024safety,alshiekh2018safe}; \\
System delay~\cite{antonova2017reinforcement,firoiu2018human}; \\
Execution uncertainty~\cite{tan2020robustifying}.
\end{tabular}
& 
\begin{tabular}[c]{@{}l@{}}
\textbf{Action space shift}: overlapping and noisy functional landscapes. \\
\textbf{Action perturbations}: injecting same-name distractor tools (with empty/incorrect parameters) \\ 
and redundant similar tools to force complex disambiguation~\cite{rabinovich2025robustness}. \\
\end{tabular}
\\
\midrule
Transition
& 
\begin{tabular}[c]{@{}l@{}}
Dynamics parameter mismatch~\cite{peng2018sim}; \\ 
Systematic dynamics bias~\cite{hanna2017grounded,da2023uncertainty}; \\
Bounded transition shift~\cite{smirnova2019distributionally}
\end{tabular}
& 
\begin{tabular}[c]{@{}l@{}}
\textbf{Transition realism}: Transient 
timeouts and partial API failures / incomplete JSON payloads~\cite{zhou2025shielda}. \\
\textbf{Transition perturbations}: modifying execution dynamics via configurable timeout rates \\
and unpredictable retry patterns to mandate autonomous recovery sequences. \\
\end{tabular}
\\
\midrule
Reward
& 

\begin{tabular}[c]{@{}l@{}}
reward signal deficiency~\cite{li2023mind,guo2024rewardaug}
\end{tabular}
& 
\begin{tabular}[c]{@{}l@{}}
\textbf{Reward realism}: Latency and financial
cost rather than overfitting to binary accuracy~\cite{kim2026cost}. \\
\textbf{Reward perturbations}: Redundant tools, misleading tool names, and hidden usage fees. \\
\end{tabular}
\\
\bottomrule
\end{tabular}
\end{table*}

\vspace{-1mm}
\section{Gaps on FM-Controlled Agents and Benchmarking Paradigm}

In Section 2, we introduced sim-to-real gaps across $\mathcal{O}$,$\mathcal{A}$,$\mathcal{T}$,$\mathcal{R}$. In this section, we further analyze how these gaps manifest in FM-controlled agents, with Table~\ref{tab:traditional-fm-connection} mapping traditional sim-to-real gap sources in RL and robotics to their analogues in FM agent settings. Moreover, benchmarking sim-to-real robustness for FM agents can draw on classical mitigation strategies from RL. 

\vspace{-1mm}
\subsection{Observation: Randomization and Shift}

FM agents possess an observation space consisting of processed textual inputs. These inputs include user queries along with tool schemas and environment feedback~\cite{li2023camel,huang2024understanding}. An observation gap emerges when the structured inputs found in simulated training benchmarks diverge from the noisy observations encountered during real-world deployment. In practice, this gap manifests through everyday user typos and massive irrelevant context distractors~\cite{liu2025evaluating} as well as two deeper structural challenges, namely multilingual misalignment~\cite{luo2025multilingual} and multi-modal noise. Multilingual misalignment describes the discrepancy where an agent fails when agents meet non-standard language beyond English. Furthermore multi-modal noise represents the critical shift from clean text observations to complex environments such as imperfect speech transcriptions or noisy optical character recognition outputs that corrupt the underlying semantic meaning~\cite{yu2024lang4sim2real,chen2025vision}.

A rigorous benchmarking paradigm should systematically expose these observation vulnerabilities through controlled randomization and distribution shift methodologies. Evaluators should inject noise into the text observations while keeping the underlying correct action unchanged. This process involves introducing minor typographical errors and query paraphrasing that mimic real-world user interactions. Furthermore distribution shift protocols must evaluate policies against entirely novel environments, such as replacing standard English instructions with diverse multilingual user inputs or previously unseen tool structures~\cite{kim2026beyond}, without altering the core functional task.

\vspace{-1mm}
\subsection{Action: Randomization and Distractors}

For FM agents, the action space translates directly to the repertoire of accessible tools or application programming interfaces. An action gap manifests when the perfectly distinct tool sets of training environments are replaced by the overlapping and noisy functional landscapes of real deployment environments~\cite{jiang2025verltool}.

To show this specific vulnerability, evaluators should systematically expand and confuse the available action space to force complex disambiguation. This process involves injecting distractor tools that share the exact name as the ground truth tool but contain empty or incorrect parameter descriptions. Evaluators can additionally introduce redundant and highly similar tools to clutter the selection pool. Perturbing the presented action set in this manner tests whether an agent relies excessively on surface cues like tool names rather than demonstrating deep semantic understanding.

\vspace{-1mm}
\subsection{Transition: Realism and Robustness}

The transition function for FM agents corresponds to the environment response following a tool execution. A transition gap emerges because agents trained under idealized assumptions expect immediate and perfect success while real deployment environments present stochastic dynamics with transient timeouts or partial application programming interface failures~\cite{zhou2025shielda}.

To expose this inherent vulnerability, researchers should subject agents to modified execution dynamics that systematically vary transition fidelity to mimic actual production environments. An effective evaluation methodology should inject configurable perturbations, such as varying timeout rates and unpredictable retry patterns, alongside incomplete data payloads like partial JSON responses. For instance, forcing an initial tool call to return a transient error mandates that the agent interprets the failure correctly and autonomously initiates a recovery sequence. Exposing models to these realistic transition conditions reveals whether policies inherently possess grounded behavior analogous to robust targets~\cite{chen2026conformal} and whether they can plan under uncertainty~\cite{liu2025uncertainty,da2024llm}.

\vspace{-1mm}
\subsection{Reward: Realism and Metadata}

A critical vulnerability of FM-controlled agents arises because agents typically overfit to accuracy and ignore underlying operational constraints such as latency~\cite{kim2026cost}. The reward gap emerges when the simulated environment fails to capture these true operational costs of real-world deployment. Furthermore, empirical observations from recent tool-use perturbation frameworks demonstrate that reward conditions consistently induce the most severe performance degradation~\cite{vuddanti2025paladin}.

To systematically benchmark, evaluators should construct scenarios where redundant tools are deliberately provided so the agent is explicitly forced to evaluate cost and latency metadata to select the optimal path.  Beyond simple redundancy, benchmarking protocols should incorporate misleading tool names and hidden usage fees to rigorously test if the model blindly optimizes for surface-level functional relevance instead of the true underlying utility. This approach effectively shifts the evaluation metric from basic semantic alignment to holistic operational efficiency, ensuring that policies learn to maximize actual deployment goals.

\vspace{-1mm}
\subsection{A Canonical Example: Observation Gap}
\label{sec:observation-example} 
Action, transition, and reward gaps are covered above and in Table~\ref{tab:traditional-fm-connection}. Here we zoom in on one \emph{observation}-gap setting: the \emph{language--execution boundary}. Standard tool-calling benchmarks~\cite{patilberkeley,chen2024t} assume predominantly English user queries and language-consistent execution (e.g., parameter values that match API conventions). In reality, users may issue queries in many languages while tool interfaces (e.g., function names, parameter identifiers, or expected value formats) remain language-invariant (e.g., English-only). Multilingual tool-calling studies~\cite{luo2025multilingual} show that models often produce \emph{semantically correct} tool calls---correct intent, correct tool selection---yet \emph{operationally invalid} ones, because parameter values are copied from the user's language (e.g., Chinese or Hindi) and violate execution-level conventions. For example, \cite{luo2025multilingual} shows that the error rate increases from 13.5\% to 28.5\% and from  5.5\% to 46.5\% when transferring the instruction from English to Chinese using GPT5~\cite{singh2025openai} and Qwen-Next-80B~\cite{qwenteam2025qwen3next} as shown in Table~\ref{tab:results}. This failure mode is called \emph{parameter value language mismatch}: the observation (user query in language L) leads the model to fill parameters with values in L, but the execution environment expects values in a fixed convention (e.g., English). The gap is thus in the \emph{observation space}---what the agent observes (multilingual natural language) is drawn from a different distribution than what was emphasized in training (often English-centric benchmarks), and the mapping from observation to action (parameter values) must satisfy execution constraints that are not fully reflected in the training distribution.

\vspace{-1mm}
\section{Research Directions}

\begin{table}[t!]
\centering
\caption{Comparison of error rates for each model's agents across different languages. The results show a significant increase in error rates for non-English languages, highlighting the parameter value language mismatch.}
\label{tab:results}

\vspace{-3mm}
\begin{tabular}{lcccc}
\toprule
Model & English & Chinese & Hindi & Igbo \\
\midrule
DeepSeek V3.2~\cite{liu2025deepseek}       & 0.100 & 0.495 & 0.400 & 0.180 \\
GPT-5~\cite{singh2025openai}               & 0.135 & 0.285 & 0.255 & 0.200 \\
Llama 3.1-70B~\cite{grattafiori2024llama}       & 0.140 & 0.475 & 0.375 & 0.310 \\
Qwen3-Next-80B~\cite{qwenteam2025qwen3next}      & 0.055 & 0.465 & 0.360 & 0.320 \\
Granite 4.0-H-Small~\cite{mishra2024granite} & 0.040 & 0.405 & 0.245 & 0.190 \\
\bottomrule
\end{tabular}
\vspace{-4mm}
\end{table}

\begin{figure}
    \centering
    \includegraphics[width=0.84\linewidth]{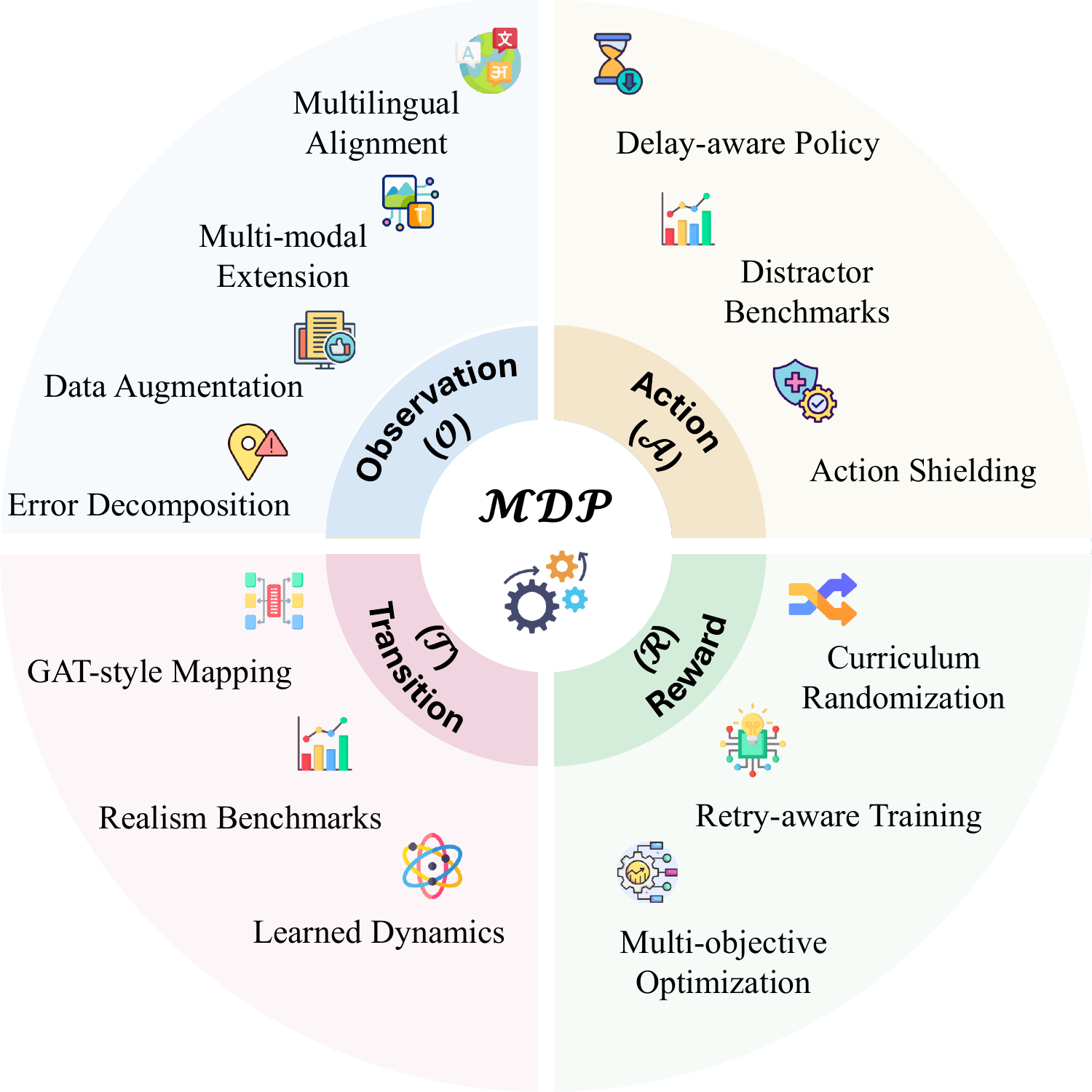}
\vspace{-3mm}
    \caption{Overview of research directions organized around the MDP loop. Each MDP component is Observation ($\mathcal{O}$), Action ($\mathcal{A}$), Transition ($\mathcal{T}$), and Reward ($\mathcal{R}$).}
    \vspace{-5mm}
    \label{fig:application}
\end{figure}

The MDP formulation above maps naturally to a set of open research directions, organized by the four components of the agent loop (Figure~\ref{fig:application}). We outline concrete opportunities for each component and close with cross-cutting challenges.

\vspace{1mm}
\noindent$\bullet$~\textbf{Observation ($\mathcal{O}$): Alignment across languages and modalities.}
The observation gap manifests most acutely in multilingual and multi-modal settings. Multilingual alignment requires training or decoding procedures that enforce parameter-value conventions across languages, so that a date format or unit convention in one language maps correctly to the same API argument as its counterpart in another. Recent multilingual agent benchmarks~\cite{hofman2025maps} reveal that non-English prompts cause disproportionate performance drops, yet few methods explicitly target the language-execution mismatch at the system level. Multi-modal extension broadens observations from text to vision, audio, and mixed-modality inputs, where sim-to-real benchmarks for vision-language agents and FM-controlled robots remain largely absent. Data augmentation, such as synthesizing paraphrases, noisy OCR outputs, or culturally varied tool descriptions, can improve robustness without additional real-world data. Error decomposition protocols that separate semantic misunderstanding from execution-level formatting errors are essential for diagnosing which part of the observation pipeline fails.

\vspace{1mm}
\noindent$\bullet$~\textbf{Action ($\mathcal{A}$): Safe and delay-tolerant execution.}
Real-world tool calls face latency, ambiguity, and risk, none of which are captured by clean-benchmark evaluation. Delay-aware policies must learn to act under variable and sometimes unpredictable tool-response times, a challenge studied extensively in real-time RL for robotics but largely unexplored for FM agents. Action shielding offers a complementary safety layer: recent work such as ShieldAgent~\cite{chen2025shieldagent} demonstrates that guardrail agents can intercept unsafe or invalid tool calls before execution through formal verification; adapting such mechanisms to the sim-to-real setting, where the simulated action space may not fully cover real API constraints, is a natural next step. Distractor benchmarks that inject irrelevant or adversarial tool options into the action space test whether agents can resist spurious choices, an analog of the distractor objects studied in robotic manipulation.

\vspace{1mm}
\noindent$\bullet$~\textbf{Transition ($\mathcal{T}$): Bridging simulated and real tool call.}
The transition function is where sim-to-real mismatch is often most severe: simulated tool calls return clean, deterministic outputs, while real APIs produce timeouts, partial failures, rate-limit errors, and non-deterministic orderings. GAT-style mapping, which adapts grounded action transformation from multi-agent settings to tool-use agents, can learn a correction function that maps simulated tool outcomes to realistic feedback distributions. Realism benchmarks should systematically vary transition fidelity by injecting configurable perturbations such as timeout rates, retry patterns, and partial JSON responses, so that robustness can be measured per perturbation type. Learned dynamics models, including LLM-based or neural surrogates that predict real-world transition distributions from simulation experience, could enable planning under transition uncertainty without costly real-world rollouts.

\vspace{1mm}
\noindent$\bullet$~\textbf{Reward ($\mathcal{R}$): Robust training under distributional shift.}
Standard RL objectives for tool-use agents overfit to clean reward signals; when deployed, noisy or delayed rewards cause the largest performance drops. Curriculum randomization, which progressively increases the severity of $\mathcal{O}$/$\mathcal{A}$/$\mathcal{T}$/$\mathcal{R}$ perturbations during training, has proven effective in robotics sim-to-real transfer and can be directly ported to agent training loops by ramping up API noise or timeout frequency over the course of training. Retry-aware training objectives that explicitly reward recovery behavior, such as successfully completing a task after an initial tool failure, encourage agents to develop robust fallback strategies rather than failing silently. Multi-objective optimization that jointly maximizes clean-task performance and stress robustness addresses the well-known tension between average-case accuracy and worst-case reliability, ensuring that hardening against perturbations does not degrade nominal performance.

\vspace{1mm}
\noindent \textbf{Cross-cutting challenges.}
Several questions span all four MDP components. It remains unknown how sim-to-real robustness scales with model size and pretraining diversity, or whether larger models naturally close $\mathcal{O}$/$\mathcal{A}$/$\mathcal{T}$/$\mathcal{R}$ gaps without targeted intervention. Shared perturbation libraries, public leaderboards reporting per-gap and combined robustness scores, and reproducibility practices such as fixed seeds, perturbation schedules, and version-controlled tool schemas are needed to reduce community fragmentation. Joint protocols that reflect real-world correlations among gap types, for example, noisy observations co-occurring with delayed transitions, and that balance coverage across languages and tool ecosystems against evaluation cost, represent the most important open benchmarking challenge.


\section{Conclusion}

Foundation model agents are rapidly transitioning from curated, failure-safe benchmarks to real-world deployments where inputs, tools, and environments are noisy and shifting. We argue that many of the resulting reliability failures are not fundamentally new but mirror the traditional sim-to-real gap studied in reinforcement learning. Therefore, we propose a unifying MDP-based framework that is structured around observation, action, transition, and reward to formalize FM agent robustness and evaluation. We hope this Blue Sky formulation catalyzes a paradigm shift in how the field diagnoses vulnerabilities, designs benchmarks, and ultimately builds reliable, real-world AI systems aligned with KDD’s goals in responsible data science and modern, dependable applications.

\begin{acks}
The work was partially supported by NSF awards \#2442477 and \#2550203. We thank the Amazon Research Awards, Cisco Faculty Research Awards, and Toyota Faculty Research Awards.
The views and conclusions in this paper are those of the authors and should not be interpreted as representing any funding agencies.
\end{acks}

\bibliographystyle{ACM-Reference-Format}
\bibliography{references}

@article{da2025survey,
  title={A survey of sim-to-real methods in rl: Progress, prospects and challenges with foundation models},
  author={Da, Longchao and Turnau, Justin and Kutralingam, Thirulogasankar Pranav and Velasquez, Alvaro and Shakarian, Paulo and Wei, Hua},
  journal={arXiv preprint arXiv:2502.13187},
  year={2025}
}

@article{zhou2026simulation,
  title={When Simulation Lies: A Sim-to-Real Benchmark and Domain-Randomized RL Recipe for Tool-Use Agents},
  author={Zhou, Xiaolin and Yuan, Aojie and Luo, Zheng and Ling, Zipeng and Pan, Xixiao and Gao, Yicheng and Zhang, Haiyue and Li, Jiate and Jiang, Shuli and Wang, Prince Zizhuang and others},
  journal={arXiv preprint arXiv:2605.11928},
  year={2026}
}

@inproceedings{zhao2020sim,
  title={Sim-to-real transfer in deep reinforcement learning for robotics: a survey},
  author={Zhao, Wenshuai and Queralta, Jorge Pe{\~n}a and Westerlund, Tomi},
  booktitle={2020 IEEE symposium series on computational intelligence (SSCI)},
  pages={737--744},
  year={2020},
  organization={IEEE}
}

@book{feinberg2012handbook,
  title={Handbook of Markov decision processes: methods and applications},
  author={Feinberg, Eugene A and Shwartz, Adam},
  volume={40},
  year={2012},
  publisher={Springer Science \& Business Media}
}

@inproceedings{tobin2017domain,
  title={Domain randomization for transferring deep neural networks from simulation to the real world},
  author={Tobin, Josh and Fong, Rachel and Ray, Alex and Schneider, Jonas and Zaremba, Wojciech and Abbeel, Pieter},
  booktitle={2017 IEEE/RSJ international conference on intelligent robots and systems (IROS)},
  pages={23--30},
  year={2017},
  organization={IEEE}
}

@inproceedings{hanna2017grounded,
  title={Grounded action transformation for robot learning in simulation},
  author={Hanna, Josiah and Stone, Peter},
  booktitle={Proceedings of the AAAI Conference on Artificial Intelligence},
  volume={31},
  number={1},
  year={2017}
}

@inproceedings{desai2020stochastic,
  title={Stochastic grounded action transformation for robot learning in simulation},
  author={Desai, Siddharth and Karnan, Haresh and Hanna, Josiah P and Warnell, Garrett and Stone, Peter},
  booktitle={2020 IEEE/RSJ International Conference on Intelligent Robots and Systems (IROS)},
  pages={6106--6111},
  year={2020},
  organization={IEEE}
}

@inproceedings{karnan2020reinforced,
  title={Reinforced grounded action transformation for sim-to-real transfer},
  author={Karnan, Haresh and Desai, Siddharth and Hanna, Josiah P and Warnell, Garrett and Stone, Peter},
  booktitle={2020 IEEE/RSJ International Conference on Intelligent Robots and Systems (IROS)},
  pages={4397--4402},
  year={2020},
  organization={IEEE}
}

@article{desai2020imitation,
  title={An imitation from observation approach to transfer learning with dynamics mismatch},
  author={Desai, Siddharth and Durugkar, Ishan and Karnan, Haresh and Warnell, Garrett and Hanna, Josiah and Stone, Peter},
  journal={Advances in Neural Information Processing Systems},
  volume={33},
  pages={3917--3929},
  year={2020}
}

@inproceedings{da2024prompt,
  title={Prompt to transfer: Sim-to-real transfer for traffic signal control with prompt learning},
  author={Da, Longchao and Gao, Minquan and Mei, Hao and Wei, Hua},
  booktitle={Proceedings of the AAAI Conference on Artificial Intelligence},
  volume={38},
  number={1},
  pages={82--90},
  year={2024}
}

@inproceedings{da2023uncertainty,
  title={Uncertainty-aware grounded action transformation towards sim-to-real transfer for traffic signal control},
  author={Da, Longchao and Mei, Hao and Sharma, Romir and Wei, Hua},
  booktitle={2023 62nd IEEE Conference on Decision and Control (CDC)},
  pages={1124--1129},
  year={2023},
  organization={IEEE}
}

@inproceedings{mehta2020active,
  title={Active domain randomization},
  author={Mehta, Bhairav and Diaz, Manfred and Golemo, Florian and Pal, Christopher J and Paull, Liam},
  booktitle={Conference on Robot Learning},
  pages={1162--1176},
  year={2020},
  organization={PMLR}
}

@inproceedings{valassakis2020crossing,
  title={Crossing the gap: A deep dive into zero-shot sim-to-real transfer for dynamics},
  author={Valassakis, Eugene and Ding, Zihan and Johns, Edward},
  booktitle={2020 IEEE/RSJ International Conference on Intelligent Robots and Systems (IROS)},
  pages={5372--5379},
  year={2020},
  organization={IEEE}
}

@article{yu2024lang4sim2real,
  title={Natural language can help bridge the sim2real gap},
  author={Yu, Albert and Foote, Adeline and Mooney, Raymond and Mart{\'\i}n-Mart{\'\i}n, Roberto},
  journal={arXiv preprint arXiv:2405.10020},
  year={2024}
}

@inproceedings{patilberkeley,
  title={The berkeley function calling leaderboard (bfcl): From tool use to agentic evaluation of large language models},
  author={Patil, Shishir G and Mao, Huanzhi and Yan, Fanjia and Ji, Charlie Cheng-Jie and Suresh, Vishnu and Stoica, Ion and Gonzalez, Joseph E},
  booktitle={Forty-second International Conference on Machine Learning},
  year={2025}
}

@inproceedings{rabinovich2025robustness,
  title={On the robustness of agentic function calling},
  author={Rabinovich, Ella and Tavor, Ateret Anaby},
  booktitle={Proceedings of the 5th Workshop on Trustworthy NLP (TrustNLP 2025)},
  pages={298--304},
  year={2025}
}

@article{wang2026agentnoisebench,
  title={AgentNoiseBench: Benchmarking Robustness of Tool-Using LLM Agents Under Noisy Condition},
  author={Wang, Ruipeng and Chen, Yuxin and Wang, Yukai and Wu, Chang and Fang, Junfeng and Cai, Xiaodong and Gu, Qi and Su, Hui and Zhang, An and Wang, Xiang and others},
  journal={arXiv preprint arXiv:2602.11348},
  year={2026}
}

@article{luo2025multilingual,
  title={Lost in Execution: On the Multilingual Robustness of Tool Calling in Large Language Models},
  author={Luo, Zheng and Kutralingam, T Pranav and Okoani, Ogochukwu N and Xu, Wanpeng and Wei, Hua and Hu, Xiyang},
  journal={arXiv preprint arXiv:2601.05366},
  year={2026}
}

@inproceedings{chen2024t,
  title={T-eval: Evaluating the tool utilization capability of large language models step by step},
  author={Chen, Zehui and Du, Weihua and Zhang, Wenwei and Liu, Kuikun and Liu, Jiangning and Zheng, Miao and Zhuo, Jingming and Zhang, Songyang and Lin, Dahua and Chen, Kai and others},
  booktitle={Proceedings of the 62nd Annual Meeting of the Association for Computational Linguistics (Volume 1: Long Papers)},
  pages={9510--9529},
  year={2024}
}

@article{da2024llm,
  title={Llm uncertainty quantification through directional entailment graph and claim level response augmentation},
  author={Da, Longchao and Chen, Tiejin and Cheng, Lu and Wei, Hua},
  journal={arXiv preprint arXiv:2407.00994},
  year={2024}
}

@inproceedings{liu2025uncertainty,
  title={Uncertainty quantification and confidence calibration in large language models: A survey},
  author={Liu, Xiaoou and Chen, Tiejin and Da, Longchao and Chen, Chacha and Lin, Zhen and Wei, Hua},
  booktitle={Proceedings of the 31st ACM SIGKDD Conference on Knowledge Discovery and Data Mining V. 2},
  pages={6107--6117},
  year={2025}
}

@article{chen2026conformal,
  title={Conformal Feedback Alignment: Quantifying Answer-Level Reliability for Robust LLM Alignment},
  author={Chen, Tiejin and Liu, Xiaoou and Nandam, Vishnu and Liou, Kuan-Ru and Wei, Hua},
  journal={arXiv preprint arXiv:2601.17329},
  year={2026}
}

@article{li2023camel,
  title={Camel: Communicative agents for" mind" exploration of large language model society},
  author={Li, Guohao and Hammoud, Hasan and Itani, Hani and Khizbullin, Dmitrii and Ghanem, Bernard},
  journal={Advances in neural information processing systems},
  volume={36},
  pages={51991--52008},
  year={2023}
}

@article{huang2024understanding,
  title={Understanding the planning of llm agents: A survey},
  author={Huang, Xu and Liu, Weiwen and Chen, Xiaolong and Wang, Xingmei and Wang, Hao and Lian, Defu and Wang, Yasheng and Tang, Ruiming and Chen, Enhong},
  journal={arXiv preprint arXiv:2402.02716},
  year={2024}
}

@article{kim2026beyond,
  title={Beyond Perfect APIs: A Comprehensive Evaluation of LLM Agents Under Real-World API Complexity},
  author={Kim, Doyoung and Ren, Zhiwei and Hao, Jie and Sun, Zhongkai and Wang, Lichao and Ma, Xiyao and Ye, Zack and Han, Xu and Yin, Jun and Ji, Heng and others},
  journal={arXiv preprint arXiv:2601.00268},
  year={2026}
}

@inproceedings{bellemare2016increasing,
  title={Increasing the action gap: New operators for reinforcement learning},
  author={Bellemare, Marc G and Ostrovski, Georg and Guez, Arthur and Thomas, Philip and Munos, R{\'e}mi},
  booktitle={Proceedings of the AAAI conference on artificial intelligence},
  volume={30},
  number={1},
  year={2016}
}

@article{wiltzer2024action,
  title={Action gaps and advantages in continuous-time distributional reinforcement learning},
  author={Wiltzer, Harley and Bellemare, Marc and Meger, David and Shafto, Patrick and Jhaveri, Yash},
  journal={Advances in Neural Information Processing Systems},
  volume={37},
  pages={47815--47848},
  year={2024}
}

@inproceedings{tan2020robustifying,
  title={Robustifying reinforcement learning agents via action space adversarial training},
  author={Tan, Kai Liang and Esfandiari, Yasaman and Lee, Xian Yeow and Sarkar, Soumik and others},
  booktitle={2020 American control conference (ACC)},
  pages={3959--3964},
  year={2020},
  organization={IEEE}
}

@inproceedings{liu2024robust,
  title={Robust deep reinforcement learning with adaptive adversarial perturbations in action space},
  author={Liu, Qianmei and Kuang, Yufei and Wang, Jie},
  booktitle={2024 International Joint Conference on Neural Networks (IJCNN)},
  pages={1--8},
  year={2024},
  organization={IEEE}
}

@article{jiang2025verltool,
  title={Verltool: Towards holistic agentic reinforcement learning with tool use},
  author={Jiang, Dongfu and Lu, Yi and Li, Zhuofeng and Lyu, Zhiheng and Nie, Ping and Wang, Haozhe and Su, Alex and Chen, Hui and Zou, Kai and Du, Chao and others},
  journal={arXiv preprint arXiv:2509.01055},
  year={2025}
}

@article{lin2025revealing,
  title={Revealing the Challenges of Sim-to-Real Transfer in Model-Based Reinforcement Learning via Latent Space Modeling},
  author={Lin, Zhilin and Sun, Shiliang},
  journal={arXiv preprint arXiv:2506.12735},
  year={2025}
}

@article{smirnova2019distributionally,
  title={Distributionally robust reinforcement learning},
  author={Smirnova, Elena and Dohmatob, Elvis and Mary, J{\'e}r{\'e}mie},
  journal={arXiv preprint arXiv:1902.08708},
  year={2019}
}

@article{zhou2025shielda,
  title={Shielda: Structured handling of exceptions in llm-driven agentic workflows},
  author={Zhou, Jingwen and Chen, Jieshan and Lu, Qinghua and Zhao, Dehai and Zhu, Liming},
  journal={arXiv preprint arXiv:2508.07935},
  year={2025}
}

@article{li2023mind,
  title={Mind the gap: Offline policy optimization for imperfect rewards},
  author={Li, Jianxiong and Hu, Xiao and Xu, Haoran and Liu, Jingjing and Zhan, Xianyuan and Jia, Qing-Shan and Zhang, Ya-Qin},
  journal={arXiv preprint arXiv:2302.01667},
  year={2023}
}

@inproceedings{kim2026cost,
  title={The cost of dynamic reasoning: Demystifying ai agents and test-time scaling from an ai infrastructure perspective},
  author={Kim, Jiin and Shin, Byeongjun and Chung, Jinha and Rhu, Minsoo},
  booktitle={2026 IEEE International Symposium on High Performance Computer Architecture (HPCA)},
  pages={1--16},
  year={2026},
  organization={IEEE}
}

@article{vuddanti2025paladin,
  title={PALADIN: Self-Correcting Language Model Agents to Cure Tool-Failure Cases},
  author={Vuddanti, Sri Vatsa and Shah, Aarav and Chittiprolu, Satwik Kumar and Song, Tony and Dev, Sunishchal and Zhu, Kevin and Chaudhary, Maheep},
  journal={arXiv preprint arXiv:2509.25238},
  year={2025}
}

@article{singh2025openai,
  title={Openai gpt-5 system card},
  author={Singh, Aaditya and Fry, Adam and Perelman, Adam and Tart, Adam and Ganesh, Adi and El-Kishky, Ahmed and McLaughlin, Aidan and Low, Aiden and Ostrow, AJ and Ananthram, Akhila and others},
  journal={arXiv preprint arXiv:2601.03267},
  year={2025}
}

@online{qwenteam2025qwen3next,
  author  = {{QwenTeam}},
  title   = {Qwen3-Next: Towards Ultimate Training \& Inference Efficiency},
  year    = {2025},
  month   = sep,
  day     = {10},
  url     = {https://qwen.ai/blog?id=4074cca80393150c248e508aa62983f9cb7d27cd&from=research.latest-advancements-list},
  urldate = {2026-03-11}
}

@article{liu2025deepseek,
  title={Deepseek-v3. 2: Pushing the frontier of open large language models},
  author={Liu, Aixin and Mei, Aoxue and Lin, Bangcai and Xue, Bing and Wang, Bingxuan and Xu, Bingzheng and Wu, Bochao and Zhang, Bowei and Lin, Chaofan and Dong, Chen and others},
  journal={arXiv preprint arXiv:2512.02556},
  year={2025}
}

@article{grattafiori2024llama,
  title={The llama 3 herd of models},
  author={Grattafiori, Aaron and Dubey, Abhimanyu and Jauhri, Abhinav and Pandey, Abhinav and Kadian, Abhishek and Al-Dahle, Ahmad and Letman, Aiesha and Mathur, Akhil and Schelten, Alan and Vaughan, Alex and others},
  journal={arXiv preprint arXiv:2407.21783},
  year={2024}
}

@article{yao2025instructional,
  title={Instructional agents: Llm agents on automated course material generation for teaching faculties},
  author={Yao, Huaiyuan and Xu, Wanpeng and Turnau, Justin and Kellam, Nadia and Wei, Hua},
  journal={arXiv preprint arXiv:2508.19611},
  year={2025}
}

@article{kim2024survey,
  title={A survey on integration of large language models with intelligent robots},
  author={Kim, Yeseung and Kim, Dohyun and Choi, Jieun and Park, Jisang and Oh, Nayoung and Park, Daehyung},
  journal={Intelligent Service Robotics},
  volume={17},
  number={5},
  pages={1091--1107},
  year={2024},
  publisher={Springer}
}

@article{zhu2025llm,
  title={Where llm agents fail and how they can learn from failures},
  author={Zhu, Kunlun and Liu, Zijia and Li, Bingxuan and Tian, Muxin and Yang, Yingxuan and Zhang, Jiaxun and Han, Pengrui and Xie, Qipeng and Cui, Fuyang and Zhang, Weijia and others},
  journal={arXiv preprint arXiv:2509.25370},
  year={2025}
}

@article{mishra2024granite,
  title={Granite code models: A family of open foundation models for code intelligence},
  author={Mishra, Mayank and Stallone, Matt and Zhang, Gaoyuan and Shen, Yikang and Prasad, Aditya and Soria, Adriana Meza and Merler, Michele and Selvam, Parameswaran and Surendran, Saptha and Singh, Shivdeep and others},
  journal={arXiv preprint arXiv:2405.04324},
  year={2024}
}

@article{hofman2025maps,
  title={MAPS: A Multilingual Benchmark for Global Agent Performance and Security},
  author={Hofman, Omer and Brokman, Jonathan and Rachmil, Oren and Bose, Shamik and Pahuja, Vikas and Shimizu, Toshiya and Starostina, Trisha and Marchisio, Kelly and Goldfarb-Tarrant, Seraphina and Vainshtein, Roman},
  journal={arXiv preprint arXiv:2505.15935},
  year={2025}
}

@article{chen2025shieldagent,
  title={Shieldagent: Shielding agents via verifiable safety policy reasoning},
  author={Chen, Zhaorun and Kang, Mintong and Li, Bo},
  journal={arXiv preprint arXiv:2503.22738},
  year={2025}
}

@inproceedings{badnava2023potential-based,
  title={A new potential-based reward shaping for reinforcement learning agent},
  author={Badnava, Babak and Esmaeili, Mona and Mozayani, Nasser and Zarkesh-Ha, Payman},
  booktitle={2023 IEEE 13th Annual Computing and Communication Workshop and Conference (CCWC)},
  pages={01--06},
  year={2023},
  organization={IEEE}
}

@article{guo2024rewardaug,
  title={Off-dynamics reinforcement learning via domain adaptation and reward augmented imitation},
  author={Guo, Yihong and Wang, Yixuan and Shi, Yuanyuan and Xu, Pan and Liu, Anqi},
  journal={Advances in Neural Information Processing Systems},
  volume={37},
  pages={136326--136360},
  year={2024}
}

@inproceedings{velasquez2021dynamic,
  title={Dynamic automaton-guided reward shaping for monte carlo tree search},
  author={Velasquez, Alvaro and Bissey, Brett and Barak, Lior and Beckus, Andre and Alkhouri, Ismail and Melcer, Daniel and Atia, George},
  booktitle={Proceedings of the AAAI Conference on Artificial Intelligence},
  volume={35},
  number={13},
  pages={12015--12023},
  year={2021}
}

@article{liu2025evaluating,
  title={Evaluating Robustness of Large Language Models Against Multilingual Typographical Errors},
  author={Liu, Yihong and Zhao, Raoyuan and Altinger, Lena and Sch{\"u}tze, Hinrich and Hedderich, Michael A},
  journal={arXiv preprint arXiv:2510.09536},
  year={2025}
}

@inproceedings{chen2025vision,
  title={Vision Language Model Helps Private Information De-Identification in Vision Data},
  author={Chen, Tiejin and Li, Pingzhi and Zhou, Kaixiong and Chen, Tianlong and Wei, Hua},
  booktitle={Findings of the Association for Computational Linguistics: ACL 2025},
  pages={4558--4572},
  year={2025}
}

@inproceedings{bousmalis2017unsupervised,
  title={Unsupervised pixel-level domain adaptation with generative adversarial networks},
  author={Bousmalis, Konstantinos and Silberman, Nathan and Dohan, David and Erhan, Dumitru and Krishnan, Dilip},
  booktitle={Proceedings of the IEEE conference on computer vision and pattern recognition},
  pages={3722--3731},
  year={2017}
}

@inproceedings{park2021sim,
  title={Sim-to-real visual grasping via state representation learning based on combining pixel-level and feature-level domain adaptation},
  author={Park, Youngbin and Lee, Sang Hyoung and Suh, Il Hong},
  booktitle={2021 IEEE International Conference on Robotics and Automation (ICRA)},
  pages={6300--6307},
  year={2021},
  organization={IEEE}
}

@inproceedings{bohez2017sensor,
  title={Sensor fusion for robot control through deep reinforcement learning},
  author={Bohez, Steven and Verbelen, Tim and De Coninck, Elias and Vankeirsbilck, Bert and Simoens, Pieter and Dhoedt, Bart},
  booktitle={2017 IEEE/RSJ International Conference on Intelligent Robots and Systems (IROS)},
  pages={2365--2370},
  year={2017},
  organization={Ieee}
}

@inproceedings{abbas2024safety,
  title={Safety-driven deep reinforcement learning framework for cobots: A sim2real approach},
  author={Abbas, Ammar N and Mehak, Shakra and Chasparis, Georgios C and Kelleher, John D and Guilfoyle, Michael and Leva, Maria Chiara and Ramasubramanian, Aswin K},
  booktitle={2024 10th International Conference on Control, Decision and Information Technologies (CoDIT)},
  pages={2917--2923},
  year={2024},
  organization={IEEE}
}

@inproceedings{alshiekh2018safe,
  title={Safe reinforcement learning via shielding},
  author={Alshiekh, Mohammed and Bloem, Roderick and Ehlers, R{\"u}diger and K{\"o}nighofer, Bettina and Niekum, Scott and Topcu, Ufuk},
  booktitle={Proceedings of the AAAI conference on artificial intelligence},
  volume={32},
  number={1},
  year={2018}
}

@article{antonova2017reinforcement,
  title={Reinforcement learning for pivoting task},
  author={Antonova, Rika and Cruciani, Silvia and Smith, Christian and Kragic, Danica},
  journal={arXiv preprint arXiv:1703.00472},
  year={2017}
}

@article{firoiu2018human,
  title={At human speed: Deep reinforcement learning with action delay},
  author={Firoiu, Vlad and Ju, Tina and Tenenbaum, Josh},
  journal={arXiv preprint arXiv:1810.07286},
  year={2018}
}

@inproceedings{peng2018sim,
  title={Sim-to-real transfer of robotic control with dynamics randomization},
  author={Peng, Xue Bin and Andrychowicz, Marcin and Zaremba, Wojciech and Abbeel, Pieter},
  booktitle={2018 IEEE international conference on robotics and automation (ICRA)},
  pages={3803--3810},
  year={2018},
  organization={IEEE}
}

\end{document}